\documentclass[letterpaper]{article} 
\usepackage{aaai24}  
\usepackage{times}  
\usepackage{helvet}  
\usepackage{courier}  
\usepackage[hyphens]{url}  
\usepackage{graphicx} 
\usepackage{natbib}  
\usepackage{caption} 
\usepackage{newfloat}
\usepackage{listings}
\DeclareCaptionStyle{ruled}{labelfont=normalfont,labelsep=colon,strut=off} 
\usepackage{hyperref} 
\usepackage{algorithm}
\usepackage{algpseudocode}
\usepackage{amsfonts}
\usepackage{amsmath}
\usepackage{tfrupee}
\usepackage{xcolor}
\usepackage{mathtools} 
\usepackage{diagbox}

\algnewcommand\algorithmicinput{\textbf{Input:}}
\algnewcommand\algorithmicoutput{\textbf{Output:}}
\algnewcommand\Input{\item[\algorithmicinput]}%
\algnewcommand\Output{\item[\algorithmicoutput]}%
\DeclareMathOperator*{\argmax}{arg\,max}
\DeclareUnicodeCharacter{20B9}{\rupee}

\author{
    Tuxun Lu, \textsuperscript{\rm 1}
    Aviva Prins \textsuperscript{\rm 2}
}

\affiliations {
    \textsuperscript{\rm 1} Johns Hopkins University, Baltimore, MD, USA\\
    \textsuperscript{\rm 2} University of Maryland, College Park, MD, USA\\
    \textsuperscript{\rm 1} tlu32@jhu.edu,
    \textsuperscript{\rm 2} aviva@cs.umd.edu
}

\title{An Online Optimization-Based Decision Support Tool for Small Farmers in India: Learning in Non-stationary Environments}

\begin{document}

\maketitle

\begin{abstract}

Crop management decision support systems are specialized tools for farmers that reduce the riskiness of revenue streams,  especially valuable for use under the current climate changes that impact agricultural productivity. Unfortunately, small farmers in India, who could greatly benefit from these tools, do not have access to them. In this paper, we model an individual greenhouse as a Markov Decision Process (MDP) and adapt \citet{li2019online}'s \textsc{Follow the Weighted Leader} (\textsc{FWL}) online learning algorithm to offer crop planning advice. We successfully produce utility-preserving cropping pattern suggestions in simulations. When we compare against an offline planning algorithm, we achieve the same cumulative revenue with greatly reduced runtime.

\end{abstract}

\section{Introduction}
Five out of six farmers in India are classified as small or marginal, with two or fewer hectares of arable land \cite{fao2022food}. In 2013, a survey by the government of India found that two-thirds of the 100 million small farmers in India \textit{lose} money on average \cite{nsso2013survey}. \citet{mishra2008risks} attributes the increasing incidence of farmers' suicides to the ongoing agrarian crisis: farmers' livelihoods are dependent on a sector that is declining in production and profitability, and exposure to uncertain conditions increases their vulnerability. Their cultivation choices, and by extension the diversity of both staple and micro-nutrient rich produce available in the local markets, have significant impacts on the diets of their local communities \cite{pradhan2021farming}.

Small Indian farmers are aware of climate risk and the necessity to change their cropping pattern \cite{dhanya2016farmers}. However, farmers do not have access to software that supports their planning needs. Rather, farmers' adaptation strategies are mainly influenced by their observations of seasonal climate onset and by their fellow farmers' choices \cite{jha2021farmer}. One way these farmers could reduce the riskiness of their revenue streams is to gain access to a crop management decision support system \cite{Fabregas}.  

However, developing a crop management decision support tool is challenging. We believe that a joint research effort between the artificial intelligence and agronomy fields could help close this gap. Reinforcement learning (RL) is a branch of artificial intelligence that deals with sequential decision-making in uncertain or unknown environments via a learning paradigm \cite{sutton2018reinforcement}. While crop planning is well studied and RL is a vast research area, previous work at the intersection of these two fields is limited \cite{GAUTRON2022107182}. 

We are uniquely positioned to investigate this problem. We are working with Kheyti,\footnote{\url{https://www.kheyti.com/}} an Indian startup that is pioneering a ``greenhouse-in-a-box" solution that saves water via drip irrigation, reduces exposure to climate risks, and increases yields by sevenfold.
It has provided us with extensive domain expertise, data, and insights into the preferences and perspectives of local farmers. Our goal is to develop an optimization-based decision support algorithm. The algorithm should ingest a range of input data and output a suggested slate of actions.
In this paper, we will mathematically formalize the problem as a non-stationary environment. Then, we will adapt \citet{li2019online}'s \textsc{Follow the Weighted Leader} algorithm.
This work is a first step towards understanding how to develop learning algorithms for decision support in a low-resource agronomy setting.

We evaluate our algorithmic approach to our Markovian model of an individual greenhouse. The slate of crops produced in our empirical results preserves utility. In addition, sensitivity analysis with respect to the algorithm's hyperparameters gives insight into the environment's moderate non-stationary qualities (smoothing parameter $\theta$) and the stability of an online policy (discount factor $\gamma$). Finally, comparisons against its offline planning equivalent reveal approximately equivalent performance and a significant reduction in computational costs.

\section{Related Work}
Crop planning and rotation problems have received considerable attention from the operations management and agricultural economics communities \cite{dury2012models}. Generally, the literature adopts optimization techniques where they maximize an objective function in light of some constraints using linear programming \cite{SARKER2009191, 9548921, jothiprakash2011optimal}. One critical problem with their approach is that the rewards are assumed to be constant while the price of crops is highly variable in reality with respect to many factors. Therefore, an algorithm that can learn from past experience better suits the crop-planning purpose.

Sequential decision-making in uncertain environments is central to the RL paradigm and, more generally, artificial intelligence \cite{sutton2018reinforcement}. This suite of algorithmic approaches has the potential for crop management support \cite{GAUTRON2022107182}. RL has been studied for similar problems such as fertilization or water irrigation decision support \cite{overweg2021cropgym, 8367433}. Like these papers, the value of our work is to explore the specific research opportunities and challenges that face the intersection of these two disciplines.

However, reinforcement learning can be expensive to compute. For instance, \citet{ashcraft2021machine} utilized proximal policy optimization (PPO) to optimize crop yields. Similarly, \textsc{CyclesGYM} requires training agents on samples, e.g. with PPO \cite{NEURIPS2022_4a22ceaf}. \citet{FENZ2023108160} trained a Deep Q-Network RL agent to generate crop rotation sequences. These approaches are model-free, meaning that training requires many samples. Resource constraints make the adoption of these breakthroughs challenging. 

To overcome the resource limitations, we instead utilize the classic machine learning algorithm \textsc{Follow the Leader} (\textsc{FTL}). \textsc{FTL} uses a simple approach: it tracks the performance of all actions taken over all previous steps and selects the one action that has performed best (the ``leader''). In crop planning, fully exploring the state space by taking actions to plant crops is unrealistic. Therefore, we consider a variant that more efficiently explores the state space to identify an expert action \cite{li2019online}. 

\section{Methods}
In this section, we formalize our methodological approach. First, we provide background information about the inputs to our system. We then outline the components of our model. Finally, we introduce our variant of \citet{li2019online}'s \textsc{Follow the Weighted Leader} algorithm to solve the system for a policy of actions.

\subsection{Data Sources}
The farmer and greenhouse data are provided by Kheyti. 
At present, these farmers have eight crop options: beetroot, bottle brinjal, cabbage, cauliflower, cucumber, French beans, green capsicum, and tomato. For each crop, our partners in India have provided to us with crop calendars, seasonality data, and average harvest yields (in kilograms, assuming a standard $361 \text{m}^2$ greenhouse). Some crops may be harvested multiple times. Crops belong to families; crops of the same family may not be planted back to back in order to prevent soil-borne pests and diseases.

The farmers have no storage capacity. At harvest time, the farmers take their crops to market. Wholesale market prices are published by the Directorate of Marketing and Inspection, Ministry of Agriculture and Farmers Welfare.\footnote{\url{https://agmarknet.gov.in/}} The directorate reports the daily price (\rupee/kg) and quantity (kg) of crop arrivals to each market in the country.

\subsection{Markovian Model}
\label{sec:model}
Given the agricultural and economic data, we formulate a Markov Decision Process (MDP) model $(\mathcal{S}, \mathcal{A}, P_t, R_t)$. At any moment, the state of the greenhouse $s$ is an element of the state space $\mathcal{S}$. Sequentially, an agent (farmer) takes an action $a \in \mathcal{A}$, which transitions the MDP to state $s'\in \mathcal{S}$ according to the transition function $P_t(s, a, s')$. $R_t(s, a)$ is a reward function to the agent if it takes action $a$ in state $s$. While the agent has the complete information for states, actions, and transitions, they only have access to market prices up to the previous timestep.

\paragraph{State space}
The state of a greenhouse, $s \in \mathcal{S}$, is defined by the tuple $(\texttt{crop}, \texttt{maturity}, \texttt{expiry}, \texttt{flag})$. $\texttt{crop} \in \mathcal{C}$ represents the current crop, e.g. tomatoes. $\texttt{maturity} \in \{0, 1, 2, \dots, \texttt{max\_maturity}\}$ represents how mature the crop is, from freshly-planted ($1$) to harvestable ($\texttt{max\_maturity}$). 0 represents a dead state. Similarly, $\texttt{expiry} \in \{0, 1, 2, \dots, \texttt{lifespan}\}$. A newly-planted crop starts at \texttt{lifespan} and decrements until death ($0$). Finally, we introduce the auxiliary Boolean variable \texttt{flag} in order to denote a constraint violation.

\paragraph{Action space}
Let us define the action space $\mathcal{A} \coloneqq \{\texttt{no\_act}, ~\texttt{harvest}\} \cup \{\texttt{plant}(c)~|~\forall c \in \mathcal{C}\}$. \texttt{no\_act} represents the instruction to tend to the current state of the greenhouse. The \texttt{harvest} action attempts to harvest the current crop and sell it at the current market price, without removing the crop from the ground. Finally, \texttt{plant}$(c)$ removes whatever is in the greenhouse and plants the crop $c$. 

\paragraph{Transition function}
Our transition function $P_t: \mathcal{S} \times \mathcal{A} \times \mathcal{S} \to \{0,1\}$ is deterministic. If the action is \texttt{no\_act}, the \texttt{crop} stays the same, \texttt{maturity} increments by one (up to a ceiling of \texttt{max\_maturity}), and \texttt{expiry} decrements by one (down to a floor of 0). If the action is \texttt{plant}($c$), \texttt{crop} resets to $c$ with $\texttt{maturity}=1$ and $\texttt{expiry}=\texttt{lifetime}$. If the action is \texttt{harvest}, the maturity resets such that the crop will be harvestable in \texttt{harvest\_frequency} timesteps (in the case of a repeat-harvest crop), or to a value that will not reach the harvestable state before the crop dies. 

There are some cases that will flip the constraint satisfaction \texttt{flag} bit. If crops $c$ and $c'$ belong to the same family, attempting to replace $c$ with $c'$ will yield a constraint violation. Harvesting a dead or immature crop is also a violation of constraints. In the immature case, \texttt{maturity} will increment. In the death case, \texttt{maturity} is set to zero. An out-of-season crop transitions to a dead state. In addition, attempting to plant a crop that will be out of season before it is harvestable is a constraint violation. The violation of constraints is memoryless: $\texttt{flag}=\texttt{False}$ does not persist between timesteps.

The seasonality aspect of constraint satisfaction (that crops may be in or out of season at any given timestep) is what makes the transition function $P_t$ temporal. Sometimes the combination $(s, a)$ will transition to an $s'$ with $\texttt{flag}=\texttt{True}$, and sometimes to $\texttt{flag}=\texttt{False}$.

\paragraph{Reward function}
The reward function is as follows:
\begin{equation*}
    R_t(s, a) \coloneqq
\begin{cases}
    k      & \text{if action yields a constraint violation}\\
    y_t(s.crop) & \text{if action is \texttt{harvest}}\\
    0         & \text{otherwise}
\end{cases}
\end{equation*}
$k\ll 0$ is a configurable constant to penalize constraint violations. When a farmer harvests a mature crop, they immediately go to market and gain revenue based on the current market price (recall that the farmers have no access to storage systems at the present time). $y$ represents a function that outputs the revenue of a given crop under current market prices. At any given timestep $t$, $y_t$ is unknown to the agent. The reward function is sparse and there may be considerable variation in $R$ between timesteps. 

\paragraph{Objective}
At present, we assume the objective of any individual farmer is to maximize total expected revenue over a finite horizon $T$.

Therefore, our goal is to find a policy of actions $\pi: \mathcal{S} \to \mathcal{A}$ such that under this policy $\pi^*$, the expected sum of reward is maximized:
\begin{equation}
    \pi^* = \displaystyle\argmax_\pi \mathbb{E}\left[\sum_{t=0}^T \gamma^t R_t(s_t, a_t)|\pi\right]
\end{equation}
$\gamma \in [0,1]$ is a discount factor to weigh the importance of future rewards relative to immediate rewards.

\subsection{Algorithmic Approach}
Our goal is to solve a Markov Decision Process in a non-stationary environment, as both the transitions and rewards change over time. To do so, we adapt \citet{li2019online}'s \textsc{Follow the Weighted Leader} (\textsc{FWL}) algorithm. Since the true reward matrix $R_t$ is unknown at time $t$, FWL approximates it with a weighted average of historical rewards (a method also known as exponential smoothing). In our case, $P_t$ is also updated at each timestep. See Algorithm \ref{alg:FWL} for the pseudocode of their algorithm, along with our modification. 

\begin{algorithm}[H]
    \caption{\textsc{FWL} with time-varying transition function $P_t$}
    \label{alg:FWL}
    \begin{algorithmic}[1]
        \Input Smoothing parameter $\theta \in [0, 1)$, initial state $s_0$, transition matrices $\{P_t\}$
        \item[\textbf{Initialization:}] $\hat{R}_0 \gets R_{-1}$
        \For{$t=1:T$}
            \State Update the weighted average of historical rewards:
            \begin{equation*}
                \hat{R}_t = (1-\theta)\hat{R}_{t-1}+\theta R_{t-1}
            \end{equation*}
            \State Solve the MDP $(S, A, P_t, \hat{R}_t)$ for a policy:
            \begin{equation*}
                \pi_t \in \argmax_{\pi} g_{\hat{R}_{t}}(\pi)
            \end{equation*}
            \State Execute $\pi_t$ to transition from $s_{t-1}$ to $s_t$
        \EndFor
        \Output $\pi_{t}$ at each timestep $t\in \{1, \dots, T\}$
    \end{algorithmic}
\end{algorithm}

The smoothing parameter $\theta$ governs how much weight is placed on historical rewards. A larger value adjusts the reward approximation more rapidly to a new environment, which is important in a case such as ours with highly variable market prices. While the exact prices fluctuate, the \textit{relative} price differences between crops do not---the profitability of one crop relative to another is stable. We could in the future replace the estimate of $\hat{R}_t$ with a different method that incorporates (for example) seasonality for a better approximation. 

\begin{figure*}[h]
    \centering
    \includegraphics[width=\textwidth]{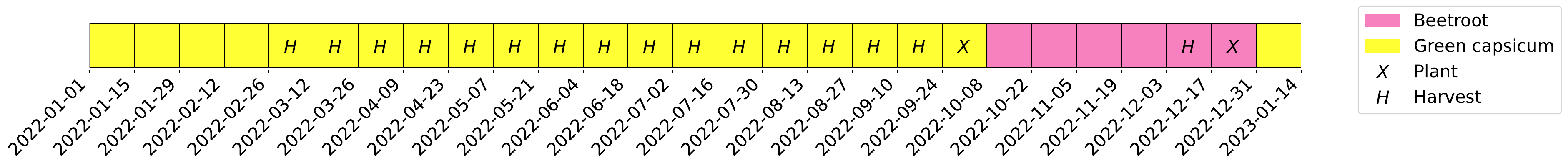}
    \caption{Example progression of states and actions yielded by our simulations. This policy produces high revenue.}
    \label{fig:policy}
\end{figure*}

To solve the MDP, we use the value iteration algorithm and the Bellman equation (Equation \ref{eq:bellman}):
\begin{equation}
    \label{eq:bellman}
    V(s) = \max_a\sum_{s'}P(s,a,s')\left[R(s,a)+\gamma V(s')\right]
\end{equation}

We reformulate the value iteration algorithm into a linear programming problem:
\begin{equation}
    \label{eq:LP}
    \begin{aligned}
        \max\quad           & \sum_{s}V(s)  \\
        \text{s.t.\quad}    & \forall s \in \mathcal{S}, \forall a \in \mathcal{A}: \\
                            & V(s) \ge \sum_{s'}P(s,a,s')\left[R(s,a)+\gamma V(s')\right]
    \end{aligned}
\end{equation}
In this formulation, the optimization problem aims to find the optimal value function that satisfies the Bellman equation and the associated MDP constraints.

\section{Experimental Setup}

We empirically evaluate the performance of the FWL algorithm (Algorithm \ref{alg:FWL}) on the MDP described in Section \ref{sec:model}. All simulations simulate the greenhouse of one ``experienced" farmer in Telangana containing one randomly newly planted crop. Unless otherwise noted, we set the smoothing parameter $\theta=0.5$, the discount factor $\gamma =0.95$, and the constraint violation penalty $k=-1e5$. Each simulation starts on January 1, 2022, and runs until December 31, 2022, with 14 days per timestep $t$. Thus, the horizon $T=26$. We report results over 10 independent trials. We use the same pseudo-random seeds in order to facilitate comparisons between simulations. 

We evaluate the results of our simulations both qualitatively and on three qualities: dynamic regret, cumulative revenue $\left(\sum_t y_t(s.crop | \pi_t)\right)$, and runtime. Dynamic regret is defined as the difference in the total reward between $\pi$ and the optimal offline policy $\pi^*$ \cite{li2019online}:
\begin{equation}
    \left\lVert \mathbb{E}\left[\sum_{t=0}^T R_t(s_t, a_t)|\pi^*\right] - \mathbb{E}\left[\sum_{t=0}^T R_t(s_t, a_t)|\pi\right] \right\rVert_\infty
\end{equation}

The simulations are executed on two laptops without integrated or external GPUs. The laptop that runs all reported runtime values operates MacOS 14.1.0 with 16 GB of RAM and an M1 Pro computer processor. The other computer is a Windows machine with 16 GB of RAM and a 12th-generation Intel i7 CPU. We implement the code in Python 3.9 and use the optimization software Gurobi v10.0.1. 

\section{Results}
\subsection{Follow the Weighted Leader Performance}

First, we investigate the performance of our algorithmic approach. To illustrate a typical output, let us look at the policy produced by a simulation with smoothing parameter $\theta=0.5$ and discount factor $\gamma=0.95$ (Figure \ref{fig:policy}). The greenhouse starts with an initial state $s_0$ of a newly planted green capsicum. The policy produced by \textsc{FWL} is to tend to and then fully harvest green capsicum (a repeat-harvest crop). Then, the policy instructs the farmer to plant and harvest beetroot. Finally, the policy plants green capsicum and the simulation ends. Relative to the other crops, beetroot is highly profitable. Recall, however, that our constraints preclude \textit{only} planting beetroot. The cumulative revenue of this policy is \rupee2,04,733 (2 lakh rupees). 

\begin{figure}[h]
    \centering
    \includegraphics[width=\linewidth]{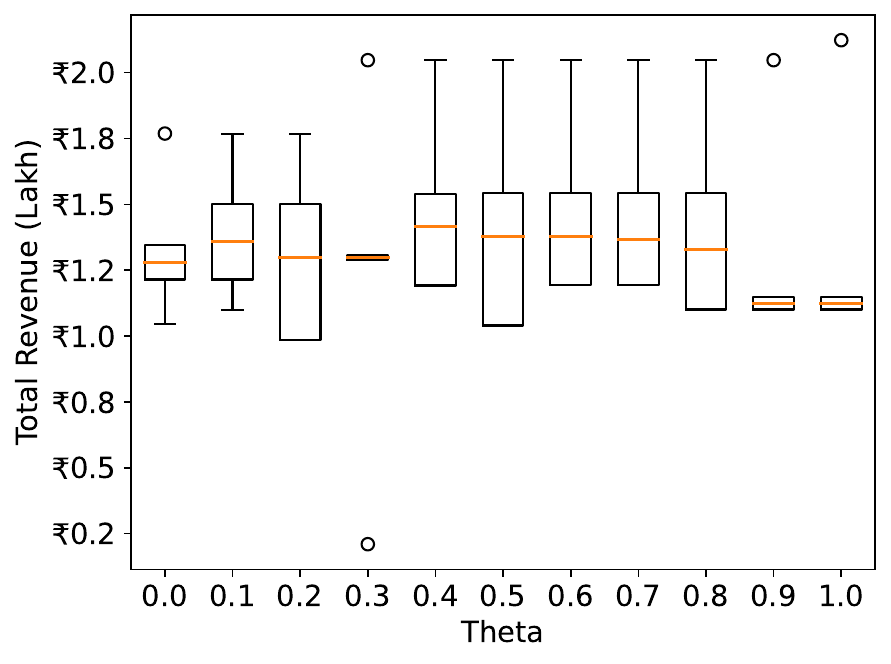}
    \caption{Middle values of $\theta$ yield the highest total revenues.}
    \label{fig:theta}
\end{figure}

The smoothing parameter $\theta$, which governs the weight of past reward observations in the weighted average calculation of $\hat{R}_t$ at each timestep, is a key parameter choice of \textsc{FWL}. We investigate the cumulative revenue yielded by each policy over multiple values of $\theta$ (Figure \ref{fig:theta}). With the same pseudo-random seed, all $\theta \in [0.3, 0.9]$ produce the previously discussed (and highly profitable) green capsicum - beetroot - green capsicum policy. At $\theta = 1.0$, the policy changes slightly to prematurely remove the green capsicum in order for the beetroot to be harvestable at an earlier and more profitable timestep. For $\theta < 0.3$, the policy plants less-profitable cucumber instead of beetroot. This example illustrates the importance of choosing a large $\theta$ in order to adapt the policy because of changes in market prices. However, it is possible to be \textit{over} receptive to market price fluctuations if the smoothing parameter is too large ($\theta \geq 0.9$, Figure \ref{fig:theta}).

There is one simulation at $\theta=0.3$ that produces unusually low total revenue. In that simulation, the policy instructions are to fully harvest the crop, and then delay planting beetroot in order to time the market. Then, the policy instructs the farmer to plant cucumbers and then immediately remove them in order to plant beetroot again, thus side-stepping the crop rotation constraint. The delay in planting beetroot and the fact that the simulation ends before the second beetroot harvest contribute to the relatively low cumulative reward of this simulation. For farmers, it is risky to rely on one high-revenue harvest.

\subsection{Learning versus Planning}

\begin{figure}[h]
    \centering
    \includegraphics[width=\linewidth]{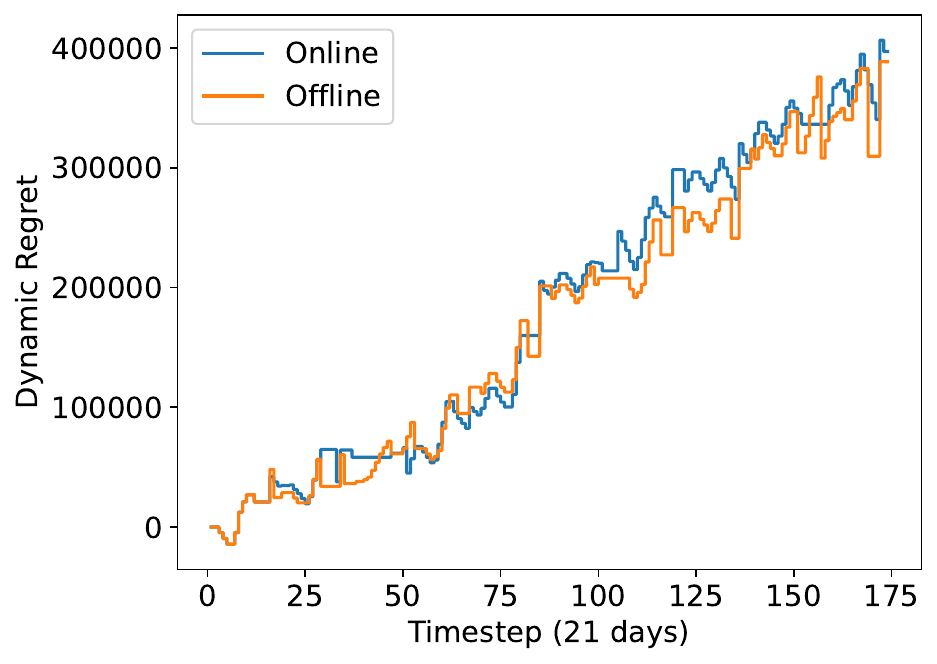}
    \caption{The dynamic regret of our online algorithm is equivalent to its analogous offline variant.}
    \label{fig:regret}
\end{figure}

\begin{figure*}[t!]
    \centering
    \includegraphics[width=\textwidth]{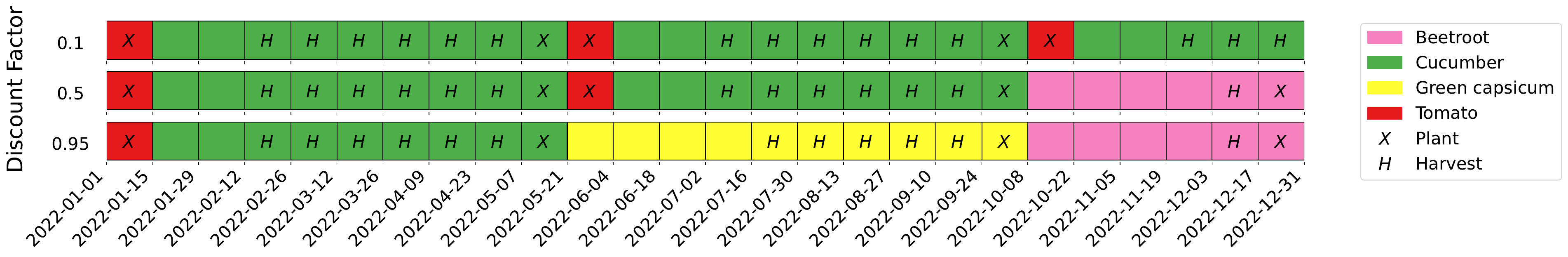}
    \caption{Progression of states and actions yielded by changing the discount factor $\gamma$. The final plant actions are green capsicum.}
    \label{fig:discount_factor_policies}
\end{figure*}

In this section, we compare the online learning algorithm against its offline planning equivalent, first proposed by \citet{AvivaPrins}. The offline variant solves the MDP specified above, with one crucial change: each combination of state and timestep $t\in \{1,~2, \dots, T\}$ is an element of the state space. Then, the transition function $P$ is stationary and known to the agent. The reward function $R$ is also stationary, however, it is not known to the agent. We approximate $\hat{R}$ with a forecast of market prices using single exponential smoothing and ten years of historical price data. We compare both models against an optimal baseline that has perfect knowledge of the true reward matrix $R$. The main focus of this experiment is to evaluate the cost of learning versus planning in terms of a) dynamic regret and b) computation complexity.

Figure \ref{fig:regret} shows the dynamic regret of the online and offline outputs. Recall that a lower regret is more desirable. At the end of the simulation, the difference in regret between the two is 8,510 (approximately 2\%). Figure \ref{fig:regret} demonstrates that the regret scales equally quickly for the two. On one hand, this alleviates the concern that constantly updating the policy in the online case introduces instability. On the other hand, it demonstrates that one does not need a particularly accurate forecast of reward in order to adequately plan a policy for multiple future timesteps (recall that in both cases, the reward forecasting method is simple, with no seasonality coefficient). 

The results in Figure \ref{fig:regret} are for only one simulation with a horizon $T=174$. This is because the offline algorithm is expensive to compute. The cardinality of our state space is determined by 
\begin{equation*}
    2\lvert \mathcal{C} \rvert \times \max_{c\in \mathcal{C}}\left(\texttt{c.max\_maturity}\right) \times \max_{c\in \mathcal{C}}\left(\texttt{c.lifespan}\right).
\end{equation*}
In the offline case, the state space increases by a multiple of the finite horizon $T$. Recall that the number of timesteps to reach the maximum maturity or lifespan of a crop scales with step size. Unlike our other simulations, the number of days per timestep is 21. Thus, the number of entries in the state space (transition matrix) is 560 (3,136,000) for the online model and 97,440 (94,945,536,000) for the offline model.

\begin{figure}[h]
    \centering
    \includegraphics[width=\linewidth]{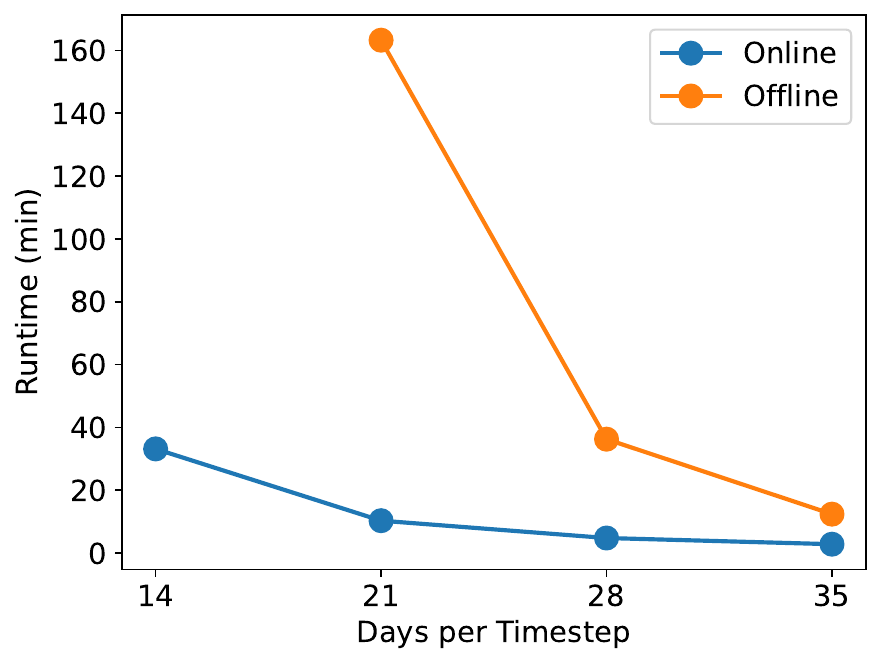}
    \caption{The runtime of the offline variant scales exponentially with more granular timestep sizes.}
    \label{fig:runtime}
\end{figure}

The space complexity of the offline variant affects its computational cost. To run the above simulations, it took two hours and 43 minutes to run the offline variant and only 10 minutes to run Algorithm \ref{alg:FWL}. Figure \ref{fig:runtime} shows a comparison of runtimes over different timestep sizes (the simulation length remains ten years). As timestep size decreases (and the granularity of the state space increases), the runtime of offline simulations increases exponentially. In contrast, the online system solves the MDP multiple times but on a smaller system. Additionally, the offline system must know the (finite) length of the horizon at planning time, while the online system can run indefinitely.

\subsection{Stability of an Online Policy}
\label{sec:discount_factor}

\begin{figure}[H]
    \centering
    \includegraphics[width=\linewidth]{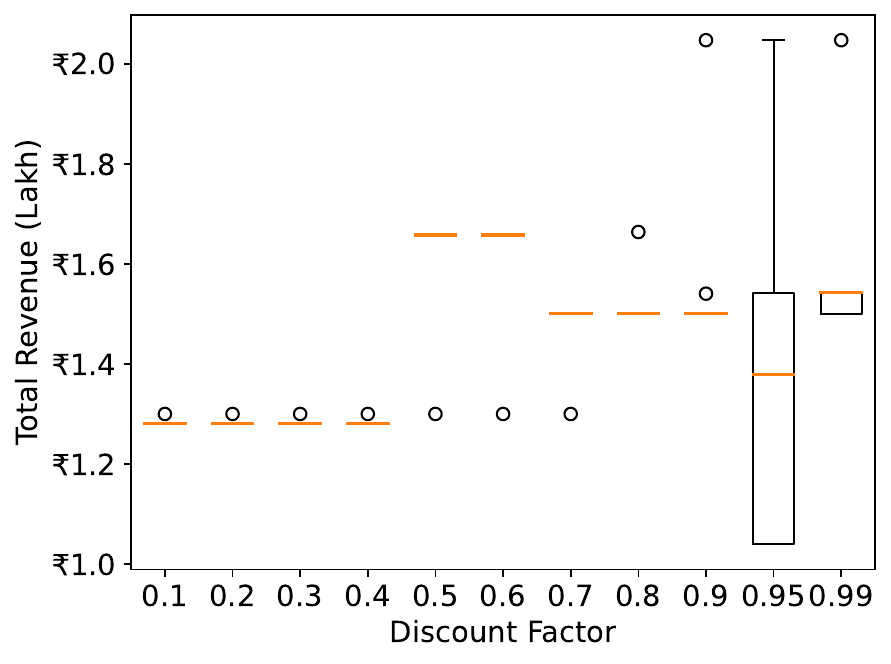}
    \caption{Immediate versus delayed reward prioritization yields different policies and cumulative revenues.}
    \label{fig:discount_factor}
\end{figure}

Finally, we discuss the impact of prioritizing long-term future rewards over greedy short-term rewards. Since the online policy updates at every timestep, it is helpful to take a more myopic view than usual (Figure \ref{fig:discount_factor}). Qualitatively, we observe that the policy is very unstable for higher values of discount factor $\gamma$. The policies $\{\pi_t\}$ output by Algorithm \ref{alg:FWL} change considerably between timesteps. At first, the policies universally instruct the farmer to override the initial state of the greenhouse by planting cucumbers or tomatoes. However, at the beginning of August, the policies override the current state of the greenhouse by planting bottle brinjal, and then override themselves again to plant green capsicum. Ironically, large values of discount factor $\gamma$ \textit{prevent} the learner from adequately maximizing future rewards because crops are replaced before they can be harvested. In contrast, lower values of discount factor $\gamma$ result in policies that are more stable but that prioritize multi-harvest crops that are generally less profitable such as cucumbers over beetroot. Ultimately, lower values of $\gamma$ result in a policy that harvests cucumbers (a lower-value crop), medium values of $\gamma$ result in a policy that successfully harvests beetroot (a high-value but risky crop), and large values of $\gamma$ eventually result in a harvest of green capsicum (Figure \ref{fig:discount_factor_policies}).

Interestingly, differences between simulations such as different initial conditions rarely result in a different policy of actions. Thus, there is a low variance in the cumulative reward for a given discount factor in Figure \ref{fig:discount_factor}. 

\section{Conclusion}
In this paper, we empirically investigate the performance of an online learning algorithm in a highly non-stationary environment. The policy of actions produced by \textsc{FWL} are intuitive, produce high cumulative revenues, and are approximately equivalent to a planning alternative. 

\paragraph{Impact} This research is the first of its kind for greenhouse-in-a-box farmers in India. The potential impact of our work is large. We are cognizant of the fact that we are working on an AI tool that may be relied upon in the future to plan out an entire group of people's livelihoods. While the potential for improvement over the status quo is great, it is vital to carefully impact the efficacy, robustness, trustworthiness, and fairness of any proposed decision support tool.

\paragraph{Future Work}

Currently, we focus on producing optimal suggestions for an \textit{individual} farmer. However, carelessly propagating this approach to all farmers will lead to poorer outcomes overall. 
In India, unbalanced supply and demand have led to extreme price fluctuations in the past, e.g., the price of tomatoes this past summer \cite{sharma_2023}. 
Farmer features, such as crop portfolios and expected yields, tend to be similar. 
Therefore, the recommendations produced by our tool would be similar. Indeed, we see evidence of this in Section \ref{sec:discount_factor}.
Local supply will greatly increase if many farmers attempt to sell their harvest at the same time.
The increased supply will cause the market price to be lower than expected. 
In future work, we will study the problem in a multi-agent setting.

There are opportunities to further refine the follow-the-leader approach for this specific decision-making problem setting. For example, we could substitute a better prediction algorithm for market prices on line 2 of Algorithm \ref{alg:FWL}. As well as analytical guarantees, we are interested in preference elicitation -- the ability to adjust $R$ to better reflect the goals of the user, such as risk reduction, portfolio diversification, or income maximization. Eventually, we would like to compare the output of our tool against a control group of farmers (i.e., conduct a randomized control study in India). 

\section{Acknowledgements}
We thank Christine Herlihy and Jasmine Stephano for helpful research discussions and code contributions. Many thanks to Dileep K H, Kaushik Kappagantulu, and the team at Kheyti for partnership and feedback, especially in developing the Markovian model. Thank you to Samuel Miller and Rebecca Morris for proofreading this manuscript. This project was partially funded by the NSF REU-CAAR grant 2150382 and NSF CAREER Award IIS-1846237.

\bibliography{main.bib}
\end{document}